# Deictic Codes, Demonstratives, and Reference: A Step Toward Solving the Grounding Problem

Athanassios Raftopoulos (raftop@ucy.ac.cy),
Department of Educational Sciences, University of Cyprus
P.O. Box 20537, 1678 Nicosia, Cyprus

Vincent C. Müller (vmueller@act.edu)
Department of Philosophy & Social Sciences, American College of Thessaloniki
P. O. Box 21021, 55510 Pylea, Greece

**Abstract**

In this paper we address the issue of grounding for experiential concepts. Given that perceptual demonstratives are a basic form of such concepts, we examine ways of fixing the referents of such demonstratives. To avoid 'encodingism', that is, relating representations to representations, we postulate that the process of reference fixing must be bottom-up and non-conceptual, so that it can break the circle of conceptual content and touch the world. For that purpose, an appropriate causal relation between representations and the world is needed. We claim that this relation is provided by spatial and object-centered attention that leads to the formation of object-files through the function of deictic acts. This entire causal process takes place at a pre-conceptual level, meeting the requirement for a solution to the grounding problem. Finally we claim that our account captures fundamental insights in Putnam's and Kripke's work on "new" reference.

## Introduction

John Campbell (1997) claims that the problem of the reference of concepts is the problem of relating concepts with imagistic content. His 'imagistic content' is the content of our experiences as we consciously access it and use it to see things as 'being such and such'. The most basic form of reference is when one perceives a thing and refers to it on the basis of one's perception by using a demonstrative, such as, "that" or "this". The reference of such a perceptual demonstrative is determined by spatial attention.

Though Campbell's thesis regarding the importance of spatial attention and perceptual demonstratives in establishing the reference of concepts to spatiotemporal objects goes in the right direction, his account has some serious problems. The main issue we wish to take on here is his view that the matter of the reference of concepts is exhausted by relating propositional with pictorial content. The problem is that both propositional and experiential content are representations, while the issue of reference is supposed to be a matter of grounding representations to the world, not of relating representations of different kinds to each other. This classical threat of infinite regress is now known under the label of 'encodingism', the view that representations are connected with the represented entities via some kind of correspondence between the two (see the critique in Bickhard, 1993 and Christiansen and Chater, 1993). Solutions that propose some such correspondence that, in turn, stands in need of an 'interpretation' cannot answer the symbol grounding problem, since they fail to account for how "symbol meaning [is] to be grounded in something other than just more meaningless symbols." (Harnad, 1990) The encoding of symbols in further symbols cannot be the solution.

We know that the problem has a solution since (some of) our concepts do have reference. A solution that does not fall into the trap of encodingism could be provided if we could single out a non-symbolic connection between our representations and the world. It seems plausible that such a connection is a causal one and that it would take place without the involvement of conceptual (i. e. symbolic) means. This is where Campbell has pointed in the right direction.

So, our task is to give a contribution to the problem how the concepts in the mind of a particular speaker can refer to objects in the world. How is it possible that I can use the concept "house" to refer successfully to certain objects? We need to single out a causal process, that could be a grounding for such reference (and that would not presuppose concepts). Once such a grounding is laid, the speaker can expand the initial grounding from experientially accessible kinds of objects to objects to which he has no such access (for contingent or principled reasons, as in the case of abstract objects). Each use of a concept would depend on such a grounding through a causal chain, starting from the initial grounding(s). How the grounding of non-experiential concepts takes place is not discussed in this paper.

In this paper we argue that spatial attention and/or object-centered attention establish the referents of certain kinds of concepts, namely perceptual demonstratives. Demonstratives are a promising start because they might rely on bodily movements in a context, not on conceptual entities that would require an interpretation. We first discuss these forms of attention and the way they individuate objects, arguing that spatiotemporal information individuates the referents and that this can be done in a bottom-up, non-conceptual way.

Then we employ Garcia-Carpintero's (2000) and Devitt's (1996) theory of demonstratives to show how the senses of demonstratives individuate their referents (demonstrata). We claim that the senses of demonstratives use the spatio-temporal information contained in the object files to fix reference. The non-conceptual use of this information provides the causal relation that grounds representations in the world. To explain how individuation takes place we employ Ballard et. al.'s (1997) theory of deictic codes and Kahneman and Treisman's (1984) theory of object-files.

Finally, in the third part of the paper, our thesis regarding reference is discussed in the context of Putnam's (1975, 1983; 1991) and Kripke's (1980) "new" theory of reference. We claim that the notion of an object-file containing predominantly spatio-temporal information provides the causal connection with the world that Putnam and Kripke sought to establish. We discuss the grounding of concepts whose referents are the basis of one's perception when one uses a demonstrative. In this sense the solution provided here, even if successful, is only the first step toward solving the problem of concept grounding in general.

**Individuating Objects**

Campbell (1997) argues that object individuation takes place by means of selective spatial attention that picks out objects features, forms feature maps, and integrates those that are found at the same location into forming objects in the way described by Treisman's *Feature Integration Theory* (*FIT*). In vision, information from different feature maps is bound together by extracting the location encoded implicitly in any feature information. Spatial attention makes the implicit location explicit. Information localized at the same location is bound together and thought to pertain to a certain object that occupies that space.

*FIT* belongs to a family of theories that hold that when one attends to an object then one automatically encodes all of its features in visual working memory. Against this, there is evidence for the existence of object-based attention which overrides featural information (other than spatiotemporal information) and which on certain occasions may pick out objects without any regard even for spatial information (Scholl & Leslie, 2000; Scholl, 2001). The role of object centered attention is primarily the parsing of a scene into discrete persisting objects, and the selection of some among these objects. The same evidence suggests that selection based on spatio-temporal information occurs very early in information processing (though segmentations of a scene into various discrete objects probably occurs at all levels of vision); in the case of vision it takes place in mid-level vision. Mid-level vision is bottom-up and cognitively impenetrable (Pylyshyn 1999; Raftopoulos, 2001), i. e. not accessible to conscious cognition, so it is not conceptual. In other words, some form of the selection of objects and the parsing of a scene is a bottom-up, cognitively impenetrable process (Carey and Xu, 2001; Scholl, 2001). Such a process would be a good candidate for the causal process we are looking for.

Given the pre-conscious processes in mid-level vision, we need to distinguish two steps: 1) *object individuation*, the processes that selects objects as discrete entities that persist in time, and 2) *object identification*, the processes that lead to the representation of objects under a certain description. The latter involves a strong semantical component, in the sense that the object represented has been identified as being such and such (e. g. a house). The former involves a much weaker level of representation. It purports to convey the sense that an object file has been opened for that specific object, that is, that the object has been "catalogued" or "indexed" as *something* that exists and persists separately of other objects with its own continuous spatio-temporal history – not as something that has certain properties (such as that of being a house). Object-files are allocated and maintained primarily on the basis of spatiotemporal information. Objects can be parsed and tracked without being identified. This representation allows access to the object but it does not describe the object. Object individuation does not require the existence of a concept associated with that object. (Of course, in theory successful object identification could also be used for individuation, as in definite descriptions like "the large house with the porch", but this presupposes grounded symbols, so we are interested in the inverse: individuation without identification.)

As an example for object individuation, think of two identical red squares that are situated in different locations. Since they are identical with regard to their features, the only way they could be treated as two distinct objects is by considering their spatiotemporal history. This presupposes that there is an object-centered attentional mechanism that is sensitive only to spatiotemporal information and not to feature information, which can pick up these objects by opening object-files. Precisely this conclusion is reached in the *MOT* (Multiple Object Tracking) experiment (Pylyshyn & Storm, 1988). In these experiments subjects must track a number of independently moving identical objects, that are initially tagged by attentional cues, among identical distractors. The success in *MOT* presupposes that the subjects attend to spatiotemporal information (relative location and direction of motion) and not to features, such as color and shape, or even the actual location of the objects. One could say that the attentional cues individuate/index in parallel the targets by assigning them tags that the subject can follow afterwards through motion. Thus, this mechanism individuates these objects and allows the subject to follow their paths and transformations while maintaining their identity as distinct objects.

Carey and Xu (2001) argue that a mechanism tracking the spatiotemporal history of objects precedes feature tracking mechanisms, and that this mechanism may override conflicting featural information. In other words, object individuation precedes feature identification – as we said above. The cognizer does not "know" or "believe" that an object moves in continuous paths, that it persists in time, though it uses this information to index and follow the object. She does not encode the object's features in some concept. She may not even have acquired the concept "object" (in this sense, you can see a house without having the concept of "house"). Object individuation may eventually result in the belief that an object is here or there, but this indexing itself does not appeal to some stored concepts regarding objects. Hence, if object individuation establishes reference, then the reference of demonstratives is not determined by a set of descriptions of features.

The discussion of object based attention shows that object files are opened and maintained on the basis of spatiotemporal information by means of cognitively impenetrable mechanisms. Petitot (1995) talks of the "positional (local) content-structure" of the scene. This positional structure is nonconceptual, and conveys information about nonvisual properties, such as causal relations (e.g., *X* "transfers" something to *Y*). In this latter category one can include the functional properties of objects, referred to as 'affordances' of objects. Suppose that one witnesses a scene in which *X* gives *Z* to *Y*. The semantics of the scene consists of two parts: (i) the semantic content of *X*, *Z*, *Y* and "give" as a specific action, and (ii) the purely positional local content. The latter is in fact the image scheme of the "transfer" type. *X*, *Y*, and *Z* occupy a specific location in the space occupied by the scene. In the image scheme, *X*, *Y*, *Z* are thus reduced to featureless objects that occupy specific relative locations, and in that sense can be viewed as pure abstract places. More specifically, *X*, *Y*, and *Z*, which in a linguistic description of the scene are the semantic roles, "are reduced to pure abstract places, locations that must be filled by 'true' participants." These places are related by means of an action, of a "transfer" type.

Petitot's "places" do not refer to the actual locations that are occupied by the objects in a scene. What Petitot seems to allude to using the spatial metaphor is the notion of an object devoid of all features (including actual location), except that it persists in time, and occupies some space – similar to the talk of "two-place predicate". It is this individuation of an object that Petitot seeks to describe by saying that the objects' only property is that they occupy their own space and this is what the notion of objects as "pure abstract places" purports to convey. In this framework, the concepts that are used in the linguistic descriptions of a scene are locational configurations, that is, spatial structures. Petitot describes the routines and algorithms of early vision that might retrieve from the morphology of a scene in a bottom-up manner the global positional information contained in it.

Several theories of mechanisms of object indexing (what we call "individuation") have been proposed. They include the *FINST* theory of visual indexing (Pylyshyn, 2001), the object-indexing theory (Scholl & Leslie, 1999), the object-files theory of Kahneman & Treisman (1984) and more recently Ballard's et. al., (1997) theory of deictic codes. The common thread of these theories is the claim that there exists a level of (visual) processing in which objects present in a scene are parsed and tracked as distinct individual objects without being recognized as particular objects that are such and such. Thus, they stress the point that object individuation precedes object identification and that there is a level of object representation that does not encode features and does not presuppose concepts; a preconceptual level of object representation. We have already discussed «object-files». We are now going to describe a plausible mechanism that allows object individuation and tracking.

**Deictic Codes, Object Individuation and Tracking**

A recent theory of deictic pointers has been developed by Ballard, et. al., (1997). They claim that at time scales of approximately 1/3 of a second, orienting movements of the body play a crucial role in cognition and form a useful computational level. At this "embodiment level," the constraints of the physical system determine the nature of cognitive operations. "The key synergy is that at time scales of about 1/3 of a second, the natural sequentiality of body movements can be matched to the natural computational economies of sequential decision systems through a system of implicit reference called deictic in which pointing movements are used to bind objects in the world to cognitive programs."

Our discussion revolves around the issue of relating internal representational states with the world. We have emphasized the potential role played by non-conceptual processes in mediating the relation between representations and the world. Now, as we mentioned, the shortest time at which bodily actions and movements, such as eye movements, hand movements, or spoken words, can be observed is the 1/3-second-time scale, the embodiment level. Thus, the mechanisms that relate conceptual content with the world through action must be sought at this level. Suppose that one looks at a scene and selects a part of it through eye focusing. The brain's representation is about, or refers to, that specific part of the scene. Acts such as the eye focusing are called "deictic strategies", from the Greek word 'deixis' (pointing at), since they are analogous to pointing with one's hand. When one's internal representation refers to an object through such a deictic representation, this is a "deictic reference."

Eye fixation exemplifies the role of deictic mechanisms, or pointers, as grounding devices, that is, as devices that ground internal representations and cognitive programs to objects in the world, through deictic reference. This binding is implemented by two functional routines in the visual system. When a scene is perceived, the eye movements perform two main functions; they extract properties of pointer locations (object identification) and they point to aspects of the environment (object localization). The second task is that of our object individuation.

**Perceptual Demonstratives and Reference**

Let us start by having a brief look at the three most influential accounts of demonstratives. The standard Fregean analysis of demonstratives considers them similar to definite descriptions and assigns them a reference (Bedeutung) and a sense (the mode of presentation of the referred object). Frege's senses are descriptive, in that they provide descriptions in terms of features of the singular term. However, demonstratives do not function quite like definite descriptions do, since demonstratives and indexicals in general are rigid designators (Kripke, 1972) whereas definite descriptions are not. A token of "that house" refers to the salient house, while "the largest house in town" may refer to one house today and to another next year. This has been used to argue that the senses of demonstratives, if any, are not descriptive. Below, we will argue that the 'senses' of demonstratives consist in causal chains that ground them in the world (a thesis similar to that of Devitt, 1996, Kripke, 1980 and Putnam, 1975; 1991). The causal chains start with a direct perceptual encounter with an object, an encounter that grounds the demonstrative in the world. Devitt (1996, 164) calls such an encounter a "grounding" – and we shall use this for the "grounding problem".

We join Garcia-Carpintero (2000), and Devitt (1996) in the view that the difference between definite descriptions and demonstratives does not discredit the role of senses of demonstratives in determining their content. Our thesis is that the content or meaning of a demonstrative consists both of its referent and its sense. This is the second main account of demonstratives. We will not argue in favor of this view, though, because it is really not crucial to the main argument developed in this paper, namely that reference construed as object individuation can be fixed by means of bottom-up perceptual processes that involve non conceptual content. What is important to this argument is the existence of such a process; this claim is independent of whether the referent is part of the meaning of demonstratives. The argument, however, as we shall see, essentially involves the role of the mode of presentation of a demonstrative in individuating objects. Thus, our claims go against the third important construal of demonstratives.

This is the direct reference theory, according to which the only content of a demonstrative is its denotation, or in other words, that the only linguistic function of a demonstrative is that it refers to its demonstratum, its referent (Kaplan, 1989). It does not have a sense. Paraphrasing Kaplan's (1989) account of the theory, one could say that a demonstrative does not describe its referent as possessing any identifying properties, it only refers to it. Though we agree that demonstratives do not provide identifying descriptions of their referent, we argue that they allow the individuation of the referent as a singular persisting object, by means of object-centered attention and spatial attention. These two provide the causal chains that ground the demonstrative. Thus, the mode of presentation of a demonstrative is not descriptive but causal. They can do that because they have a mode of presentation of the referent. But what is the mode of presentation when one says for example "that" pointing to a house?

Campbell (1997) thinks that the problem of the sense of a perceptual demonstrative is a problem about selective attention, in so far as he considers the mode of presentation to provide imagistic information related to the referent. It is the role of selective spatial attention to isolate that information in a scene that pertains to the referent. Thus, Campbell takes the mode of presentation of a demonstrative to include information that could individuate the referent on the basis of its observable features and, in an essential manner, on the basis of its spatial location. In fact, difference of location only suffice to establish difference in the mode of presentation of the same object by two different demonstratives.

Garcia-Carpintero (2000) and Devitt (1996) offer a thorough account of the senses of demonstratives, which is similar in some respects to that of Campbell's. The sense, according to Garcia-Carpintero, is an ingredient of presuppositions of acquaintance with the referent; "presuppositions" meaning "propositions that are taken for granted" when a statement is uttered. In this fashion, senses are individuating properties that allow the individuation of the referent.

Suppose one perceives something as being a house and utters the statement "that is *f*" pointing at a certain object (the house) and assigning it the property *f* (e. g. "beautiful"). The term "that" is a singular term associated with the description "the *f* house". According to Garcia-Carpintero, when one uses the singular term "that" one takes oneself to be acquainted with an object by having a 'dossier' for "the *f* house", which picks it out. The object fulfills the conditions specified in the dossier, in our case the proposition "there is a unique house most salient when the token $t$ of 'that' is produced and $t$ refers to that house." Now, the phrase "most salient when $t$ occurs" is equivalent to the expression "house in such and such a location with such and such visual features." The "in such and such a place with such and such visual features" is the mode of presentation of the token $t$ of the demonstrative "that". This mode of presentation individuates the object to which the demonstrative refers.

The dossier of the object that acquaints one with the object can be updated by new information, by adding contents or by revising its content. One notes a distinction between an object being singled out as the referent of a demonstrative and its acquaintance dossier (file). The latter ontologically presupposes the former; one needs an object to create its dossier. One also needs to ensure that the object with such and such features at time $t_1$ is the same object with such and such features at time $t_2$. Perception must provide for a mechanism that establishes the existence of an object as a distinct entity and opens a dynamic file on it. One needs, in other words, a mechanism that individuates the demonstrata of perceptual demonstratives.

### Object Individuation and Reference

Let us see where we stand with regard to the issue of the reference of perceptual demonstratives related to object individuation. When one uses a demonstrative one opens a file for the object being demonstrated. According to the psychological evidence, the first thing that this file does is to individuate the object based on spatiotemporal information. This ensures the existence of a distinct object whose paths in space and time can be tracked. The object file thus allows acquaintance with the referent of the demonstrative, and in this sense, it constitutes its mode of presentation. Kahneman's "object-file" becomes a truncated version of Garcia-Carpintero's "dossier", a dossier that contains only spatiotemporal information. As the object moves in space-time, feature detection mechanisms infuse the file with feature information allowing feature *identification* (the full "dossier".)

Let us investigate the power of the account sketched so far with Brian Loar's (1976) example, also used by Garcia-Carpintero (2000) to argue that descriptive senses fix the referents of the terms with which they are associated: Suppose that Smith and Jones see a man on the train every morning. One evening they watch a man being interviewed on a television show, they are unaware that this man is the same man they meet on the train every morning, and it so happens that during the show they have just been talking about the man on the train. Suppose now that Smith switches his attention to the man on the television and says, "he is a stockbroker", referring to the man on the television. Jones, unaware of Smith's attention switch, takes Smith to refer to the man on the train about whom they have been talking. Though Jones has correctly identified the referent, since the man on the train is the same as the man on the television, one feels that Jones has failed to understand Smith's utterance. This shows that the manner of presentation of singular terms is important even on referential uses for grasping the meaning of what is being communicated.

The upshot of Loar's example is that although Jones' belief to the effect that the man on the train is a stockbroker has the same truth conditions as Smith's belief that the man on the television is a stockbroker (since the referent in both beliefs is the same person), Smith is justified in holding his belief, whereas Jones' is not.

Jones missed the information that would have justified his belief, because he does not know that the man on the television and the man on the train are the same person. So, for Jones, and thus information pertaining to the former does not apply to the latter. To use the terminology of this paper, Jones has two different object-files; one for the person on the television and one for the person whom he meets on the train. The role of the mode of presentation of a singular term is to clarify this point, namely whether the object under consideration has been individuated the appropriate way. Spatiotemporal information purports to do exactly that: had Jones followed the spatiotemporal path of the person on the train, he would have known that it is the same person that appears on the television and he would have used all relevant information to update that person's object-file; so his belief would have been just as justified as Smith's.

It seems thus, that object individuation (the mode of presentation) is indispensable to fixing the referent of a perceptual demonstrative. The individuation is accomplished by opening an object-file fixing the object to which the demonstrative refers and allowing its tracking. In the course of tracking, additional information, e. g. on shape and color, may be added to the "dossier" to allow tracking in difficult circumstances (as when one thing is inside some other thing). It is essential for the success of concept grounding and the escape from the regress of encodingism that this individuation process is not cognitively penetrable. No conceptual content, no existing representations can be used in the individuation process, so it has to be inaccessible to conscious content-laden processing. Also, individuation should not be seen as establishing a concept – this is what happens in the step of identification. Individuation just grounds the concept, fixing it onto an object so that the concept can be "filled" with information.

### New Theories of Reference

If object individuation can fix reference and if object individuation can be carried out without conceptual involvement, then reference can be fixed in a nonconceptual manner. Of course, this goes against the standard descriptive theories of reference, according to which a sign is associated with a concept in the mind, a "sense", which constitutes its meaning and determines what the sign refers to. It allows one to pick out the objects in the environment that are 'fall under' the concept. The reference of a word is fixed by certain of the descriptions associated with the word: that thing over there counts as a "house", given that it is a building which could be used as a human dwelling.

A problem with these kinds of theories has been expressed in terms that remind strongly of the symbol grounding problem: Devitt (1996, 159) argues that descriptive theories of reference are incomplete because by explaining references by descriptions, they appeal to the application of descriptions of other words; thus, they explain reference by appealing to the reference of other words. To escape the lurking infinite regress, there must be some words whose reference does not depend on that of other words, that is words that are founded directly in the world.

Kripke (1972) and Putnam (1975; 1983; 1991) have argued that the standard conception of reference fails for certain kinds of words, namely demonstratives, proper names and natural kind terms. It is interesting to see whether our notion of reference is compatible with Putnam's (1975; 1983) direct-reference theory ("direct" in that it avoids the mediation of conceptual content in establishing reference). According to this theory, descriptions ascribing properties would identify the wrong referents of the terms. Once a causal contact between concept and object is established, the world itself has a say on the fixing, what Putnam (1991) will later call the "contribution of environment".

Putnam (1991) argues that there is an indexical (deictic) component that participates in reference fixing. When one takes a liquid sample to be water, one does so because one thinks that this liquid sample has a property, namely, "the property of behaving like any other sample of pure water from our environment" (Putnam 1991, 33). This property is not a purely qualitative property (meaning that membership is not determined by a set of criteria); its description involves a particular example of water, one given by pointing or focusing (hence, the term '"indexical"). The stuff out there, to which the act of pointing is an essential part of fixing reference of the natural kind term "water", is the contribution of the environment. Putnam says that "… the extension of certain kinds of terms … is not fixed by a set of 'criteria' laid down in advance, but is, in part, *fixed by the world*. There are *objective laws* obeyed by multiple sclerosis, by gold by horses, by electricity; and what it is rational to include in these classes will depend on what those laws turn out to be." (Putnam 1983, 71). This brings into mind the notion of causal chains by means of which demonstratives refer, causal chains that are established through object based and spatial attention.

Kripke (1980) refers to this assigning of names as "initial baptisms". Suppose, that one points to a star and says, "that is to be Alpha Centauri" (Kripke, 1980, 95). By this one commits himself to the following: "By 'Alpha Centauri' I shall mean that star over there with such and such coordinates." Kripke (1980, 135) claims that the reference of general natural kind terms is similarly fixed: "the reference (of singular terms) can be fixed in various ways. In an initial baptism ostentation or a description typically fixes it. … The same observation holds for such a general term as 'gold'."

These are telling examples, because they point out the role of spatial information and of object based attention in fixing the reference of singular terms. The causal chain that grounds the term starts with spatio-temporal non-descriptive information that opens an object-file for some object. This way of fixing the referents of singular and natural kind terms captures adequately Kripke's intuition that: "Don't ask: how can I identify this table in another possible world, except by its properties? I have the table in my hands, I can point to it, and when I ask whether *it* might have been in another room, I am talking, by definition, about *it*. I don't have to identify it after seeing it through a telescope. If I am talking about it, I am talking about *it*." (1980, 52-53). Though Kripke speaks of proper names, his analysis easily transfers to all singular terms, and thus, to perceptual demonstratives (Garcia-Carpintero, 2000). Names and indexicals, are associated with something extralinguistic, their referents. Some existentially given thing is essential in fixing these referents.

We have claimed that the mode of presentation of the referent by its demonstrative is essential in reference fixing, and we have argued that the mode of presentation fixes reference by opening an object-file for the referent of the demonstrative. This object-file includes spatiotemporal information and its function is to individuate the referent, that is, to establish the existence of a distinct body that perseveres through space and time. It establishes the causal continuity with the thing originally "pointed at" by the perceptual demonstrative, satisfying Putnam's criterion for reference fixing. It also provides the causal relation between the representation and the world that grounds the former in the latter. The object-file provides the indexical component that participates in reference fixing. The content of the object-file being retrieved in a bottom-up manner warrants that this object file is the 'contribution of the environment' and not the contribution of conceptual content.

## Conclusion

We argue that perceptual demonstratives capture the essential way in which one refers to objects in one's experience. The sense of "demonstrative reference" involved, however, departs from the notion of the referent as an object that is individuated by some description. The representation of the referent in the sense intended here does not encode any featural properties, is pre-conceptual, and the process that leads to its formation is cognitively impenetrable. The only property that the individuated referent has is that it is being tagged as a discrete object that persists in time and occupies some space, and thus, is being rendered accessible to the viewer. We claim that this process of reference fixing provides the causal relation required to solve the grounding problem.